\documentclass[lettersize,journal]{IEEEtran}
\usepackage{amsmath,amsfonts}
\usepackage{algorithmic}
\usepackage{algorithm}
\usepackage{array}
\usepackage[caption=false,font=normalsize,labelfont=sf,textfont=sf]{subfig}
\usepackage{textcomp}
\usepackage{stfloats}
\usepackage{url}
\usepackage{verbatim}
\usepackage{graphicx}
\usepackage{cite}
\usepackage{color}
\usepackage{multirow}
\usepackage{indentfirst}
\usepackage[colorlinks,
            linkcolor=red,
            anchorcolor=blue,
            citecolor=green 
            ]{hyperref}
\begin{document}

\title{CrossFuse: Learning Infrared and Visible Image Fusion by Cross-Sensor Top-K Vision Alignment and Beyond}

\author{Yukai Shi\dag, Cidan Shi\dag, Zhipeng Weng, Yin Tian, Xiaoyu Xian, Liang Lin,~\IEEEmembership{Fellow, IEEE}

\thanks{Y. Shi, C. Shi and Z. Weng are with School of Information Engineering, Guangdong University of Technology, Guangzhou, 510006, China (email: ykshi@gdut.edu.cn; cdshi923@gmail.com; 2112303036@mail2.gdut.edu.cn ).}

\thanks{Y. Tian and X. Xian are with CRRC Academy Co., Ltd., Beijing, China (e-mail: ty@crrc.tech; xxy@crrc.tech). (\emph{Corresponding author: Xiaoyu Xian})} 

\thanks{ L. Lin is with School of Computer Science, Sun Yat-sen University, Guangzhou, 510006, China. (email: { linliang@ieee.org}).}

\thanks{
 {\dag} The first two authors share equal contribution.
 }
}


\maketitle

\begin{abstract}
Infrared and visible image fusion (IVIF) is increasingly applied in critical fields such as video surveillance and autonomous driving systems. Significant progress has been made in deep learning-based fusion methods. However, these models frequently encounter out-of-distribution (OOD) scenes in real-world applications, which severely impact their performance and reliability. Therefore, addressing the challenge of OOD data is crucial for the safe deployment of these models in open-world environments. Unlike existing research, our focus is on the challenges posed by OOD data in real-world applications and on enhancing the robustness and generalization of models. In this paper, we propose an infrared-visible fusion framework based on Multi-View Augmentation. For external data augmentation, Top-k Selective Vision Alignment is employed to mitigate distribution shifts between datasets by performing RGB-wise transformations on visible images. This strategy effectively introduces augmented samples, enhancing the adaptability of the model to complex real-world scenarios. Additionally, for internal data augmentation, self-supervised learning is established using Weak-Aggressive Augmentation. This enables the model to learn more robust and general feature representations during the fusion process, thereby improving robustness and generalization. Extensive experiments demonstrate that the proposed method exhibits superior performance and robustness across various conditions and environments. Our approach significantly enhances the reliability and stability of IVIF tasks in practical applications.
\end{abstract}


\section{Introduction}
\IEEEPARstart{I}{mage} fusion integrates significant and complementary information from multi-source images to generate high-quality and valuable fused results~\cite{xu2020u2fusion, ma2022swinfusion, zhao2023tufusion, gao2022dcdr, zhou2024decoupled}. Image fusion encompasses various categories, 
including infrared and visible image fusion (IVIF)~\cite{zhang2023visible},
medical image fusion~\cite{nie2020multi, mu2023learning}, digital photography image fusion~\cite{li2022learning, qu2022transmef}, 
and remote sensing image fusion~\cite{roy2023multimodal, qiao2023boosting}.
IVIF is a prominent and challenging subfield with broad applications in pedestrian detection and recognition~\cite{braso2020learning}, video surveillance~\cite{bhatnagar2015novel, paramanandham2018infrared}, and autonomous driving~\cite{li2020ivfusenet}. Images obtained from different sensors in IVIF possess distinct but complementary characteristics. Visible images offer high spatial resolution and detailed textures, but are sensitive to lighting conditions. Infrared images capture thermal radiation information and are less affected by lighting variations, but provide lower spatial resolution and blurred texture. By integrating these complementary characteristics, fusion results highlight prominent targets and enrich texture details with enhanced contrast, thereby providing a comprehensive scene representation and facilitating advanced vision tasks~\cite{wang2021cgfnet, zhou2021ecffnet}.

Extracting essential feature representations from source images is a key step in image fusion. Early traditional methods, such as multi-scale transformation~\cite{li2011performance}, sparse representation~\cite{liu2016image, zong2017medical}, and subspace learning~\cite{xing2021hyperspectral}, achieved image fusion by identifying commonalities across different source images and manually designing fusion rules. In recent years, deep learning-based methods have made remarkable progress in image fusion, including autoencoder (AE)-based~\cite{li2018densefuse, li2021rfn}, convolutional neural network (CNN)-based~\cite{zhang2020ifcnn, zhang2021sdnet}, generative adversarial network (GAN)-based~\cite{ma2020infrared, yang2021infrared}, 
and Transformer-based~\cite{ma2022swinfusion, tang2023datfuse} architectures. Additionally, hybrid architectures such as CNN-Transformer are increasingly being integrated into image fusion~\cite{zhao2023cddfuse, jiang2024mutual}. 
Recent methods have specifically addressed challenges related to illumination and color consistency in infrared-visible fusion, further advancing the quality and robustness of fusion outcomes in complex visual tasks~\cite{tang2023divfusion, yue2023dif, chen2024lenfusion, yao2024navigating}. These deep learning-based approaches leverage the powerful feature extraction capabilities of deep neural networks, combined with finely designed loss functions, to consistently improve fusion outcomes and drive advancements in various visual applications.

Although current learning-based fusion methods have demonstrated excellent performance, their model training typically occurs in idealized, static, and closed environments. In these settings, test data originate from the same labeled and feature space distribution as the training data, known as In Distribution (ID). However, real-world applications present models with data from unknown domains or distributions, referred to as Out-of-Distribution (OOD) data~\cite{hendrycks2018benchmarking, miller2021accuracy}. This can degrade model performance and impact their safe deployment in the open-world scenarios~\cite{sehwag2019analyzing,shi2024nitedr}. 
Particularly in IVIF tasks, models are trained on specific datasets, yet practical applications experience varying environmental conditions that lead to significant shifts in data distribution. The challenge of OOD scenarios impacts model performance, robustness, and reliability in real-world applications. For instance, in the field of autonomous driving, model vulnerability to OOD data can lead to minor errors with potentially catastrophic consequences~\cite{koopman2017autonomous}. Therefore, it is imperative to develop reliable and robust fusion methods to address OOD scenarios in real-world applications, ensuring that models maintain high performance and robustness across different domains or environments.

\begin{figure*}[!t]
\centering
\includegraphics[width=0.95\linewidth]{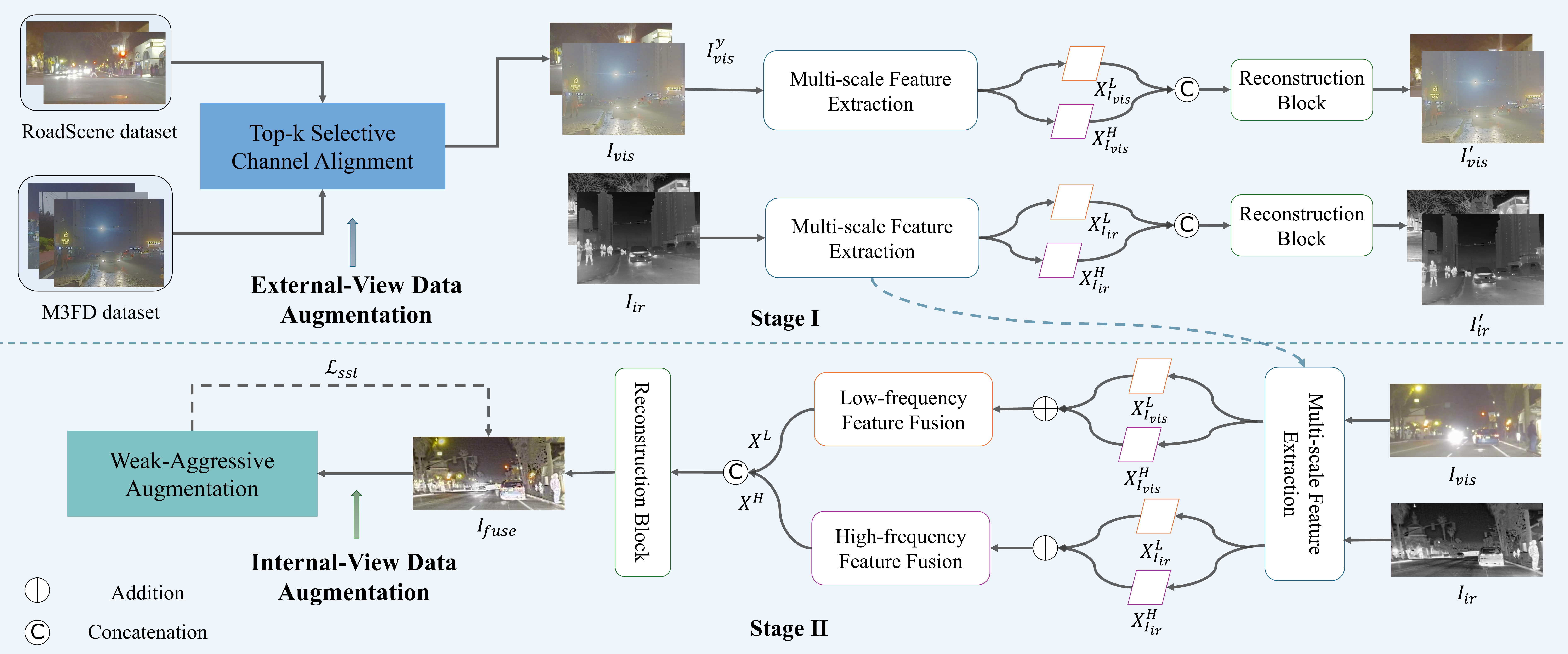}
\caption{Overall framework of the proposed method. External data augmentation utilizes Top-K selective channel alignment to effectively expand training data scale. Internal data augmentation further employs weak-aggressive augmentation for self-supervised learning. The frequency-aware fusion network achieves more comprehensive and detailed feature extraction and fusion.}
\label{fig1}
\end{figure*}

To address the challenges, we propose an innovative approach based on a data-centric strategy. Given the non-stationarity and complexity of real-world scenarios, we aim to enhance model performance through a cross-dataset learning algorithm. 
Auxiliary datasets usually have significant distribution differences from the original training data. Direct use may prevent the model from effectively capturing the original data features and lead to performance degradation. Therefore, from the external view, RGB-wise channel alignment combined with Top-K selection~\cite{li2022deep, chen2023learning, chen2024towards, xiao2024ttst} is utilized to narrow the distribution gap between external and target data, enabling the model to better adapt to unknown real-world data distributions. Furthermore, from the internal view, we introduce weak-aggressive augmentation~\cite{bai2022msr} to establish self-supervised learning. Through a specifically designed loss function, the model learns visual representations from diverse augmented views, thereby enhancing its robustness.

Specifically, we propose a novel infrared-visible fusion framework centered on multi-view data augmentation. 
This framework integrates external data augmentation (DA) using Top-k Selective Channel Alignment and internal DA for self-supervised learning, supplemented with a frequency-aware fusion network.
For the external view, Top-k Selective Channel Alignment alleviates data distribution shifts by applying selective RGB-wise gamma transformation~\cite{liu2023data}, effectively introducing augmented samples and improving the generalization capability of the model. For the internal view, weak-aggressive augmentation introduces different augmented views to facilitate self-supervised learning. The frequency-aware fusion network generates fused images containing comprehensive global structural characteristics and rich local texture details. Through this multi-view data augmentation, the model acquires more generalized and robust feature representations, enabling it to flexibly adapt to open and dynamic real-world environments.

In brief, the main contribution of this paper is as follows:
\begin{enumerate}
    \item {We explore the OOD scenario challenges of model deployment in real-world applications. Given a OOD data, our model performs cross-sensor  transformations and augmentation strategies on pre-existing datasets for generalizable feature learning.}
    \item{External-view data augmentation mitigates distribution shifts through Top-k Selective Channel Alignment, addressing distribution mismatch issues when directly utilizing external data. By performing RGB-wise transformations on diverse data distributions, the adaptability of the model is greatly improved.}
    \item{Internal-view data augmentation employs a self-supervised loss function to learn feature representations from different weak-aggressive augmentation views, improving the ability of the model to generalize across domains with limited labeled data.}
    \item{Extensive experimental results on multiple datasets demonstrate that our method achieves outstanding fusion performance across various conditions and distributions.}
\end{enumerate}

\begin{table*}[!t]
\caption{A demonstration of Out-of-distribution problem. The performance of each existing method degrades to varying degrees, especially the testset exhibits different pattern toward training samples.}
\centering
\resizebox{2\columnwidth}{!}{
    \begin{tabular}{|c|c c c|c c c|c c c|c c c|}
        \hline
        \multirow{3}{*}{Method} & \multicolumn{6}{c|}{Training Set: RoadScene} & \multicolumn{6}{c|}{Training Set: MSRS}\\ \cline{2-13}
        & \multicolumn{3}{c|}{Test on RoadScene} & \multicolumn{3}{c|}{Test on MSRS} & \multicolumn{3}{c|}{Test on MSRS} & \multicolumn{3}{c|}{Test on RoadScene}\\ \cline{2-13}
        & SF & AG & SCD & SF & AG & SCD & SF & AG & SCD & SF & AG & SCD\\
        \hline
        \hline
        Densefuse & 6.13 & 2.69 & 1.25 & 4.56(\textcolor{green}{$\downarrow$21.4\%}) & 1.68(\textcolor{green}{$\downarrow$18\%}) & 1.14(\textcolor{green}{$\downarrow$8.1\%}) & 5.8 & 2.05 & 1.24 & 8.5 & 3.46 & 1.17(\textcolor{green}{$\downarrow$6.4\%})\\
        \hline
        DIDFuse & 14.49 & 5.6 & 1.77 & 9.64(\textcolor{green}{$\downarrow$7.8\%}) & 2.01(\textcolor{green}{$\downarrow$45.8\%}) & 1.09 & 10.45 & 3.71 & 0.26 & 15.4 & 5.78 & 1.3(\textcolor{green}{$\downarrow$26.5\%})\\
        \hline
        RFN-Nest & 10.99 & 4.24 & 1.47 & 5.54(\textcolor{green}{$\downarrow$10.1\%}) & 1.89(\textcolor{green}{$\downarrow$14.5\%}) & 1.16 & 6.16 & 2.21 & 1.09 & 8.3(\textcolor{green}{$\downarrow$24.5\%}) & 3.55(\textcolor{green}{$\downarrow$16.3\%}) & 1.33(\textcolor{green}{$\downarrow$9.5\%})\\
        \hline
        PIAFusion & 18.41 & 6.53 & 0.67 & 10.24(\textcolor{green}{$\downarrow$6.2\%}) & 3.07(\textcolor{green}{$\downarrow$13.5\%}) & 0.68(\textcolor{green}{$\downarrow$56.4\%}) & 10.92 & 3.55 & 1.56 & 11.26(\textcolor{green}{$\downarrow$38.8\%}) & 4.12(\textcolor{green}{$\downarrow$36.9\%}) & 1.24\\
        \hline
        TarDAL & 12.06 & 4.59 & 1.33 & 9.91(\textcolor{green}{$\downarrow$58.5\%}) & 3.12 & 1.48 & 23.86 & 2.46 & 0.26 & 23.01 & 5.35 & 0.76(\textcolor{green}{$\downarrow$42.8\%})\\
        \hline
    \end{tabular}
}
\label{tab1}
\end{table*}

\section{Related Works}
In this section, we first review representative deep learning-based IVIF methods and then discuss the challenges posed by OOD data in real-world applications.

\subsection{Deep Learning-based IVIF}
Deep learning-based IVIF methods primarily includes those based on AE, CNN, GAN, and Transformer as well as CNN-Transformer.

AE-based methods leverage pre-trained autoencoders on large-scale datasets for feature extraction and image reconstruction. These approaches integrate deep features from different modality images through manually designed fusion rules. Li et al. proposed Densefuse~\cite{li2018densefuse}, which introduces a densely connected encoder network for feature extraction, while the corresponding decoder network produces the fusion result based on a manually crafted fusion strategy. Ren et al. developed an image fusion network based on Variational Autoencoders (VAE)~\cite{ren2021infrared}. Attention mechanisms have also been incorporated into AE-based fusion frameworks to better focus on characteristics from source images~\cite{liu2021learning}. However, the performance of these methods is constrained by the reliance on manually designed fusion rules.

In CNN-based image fusion methods~\cite{liu2020bilevel, zhang2020ifcnn, zhang2021sdnet, li2021rfn, xian2024crose}, end-to-end fusion architectures~\cite{li2021rfn, zhang2021sdnet} achieve single-step inference for feature extraction, fusion, and image reconstruction. The incorporation of various network modules and loss functions enhances feature extraction and fusion, resulting in high-quality fused images. Furthermore, in light of integrating fusion with advanced visual tasks, Tang et al. proposed SeAFusion~\cite{tang2022image} to explore multi-task learning in high-level vision. Yao et al.~\cite{yao2025color} further address the challenge of nighttime image fusion with low-light enhancement technology and knowledge distillation. Additionally, HG-LPFN~\cite{yao2023laplacian} present an novel laplacian pyramid network with hierarchical guidance to explore end-to-end IVIF. However, CNN-based architectures have limitations in extracting out-of-distribution features, posing difficulties in generating fused image with complete content.

GAN-based methods~\cite{ma2019fusiongan, fu2021image, liu2022target} address the challenge of insufficient supervision information in image fusion through adversarial learning. FusionGAN~\cite{ma2019fusiongan} first models the image fusion problem as an adversarial game between a generator and a discriminator, leveraging the distribution properties of source images to generate fused results. To maintain a balance of information from different modality, dual-discriminator structures~\cite{xu2019learning, huang2023magan} and attention mechanisms~\cite{li2020attentionfgan} have been introduced into fusion methods. However, persistent issues of training instability and interpretability limitations continue to affect these methods, resulting in artifacts and blurred details in the generated images.

The powerful modeling capability of the Transformer~\cite{vaswani2017attention} in handling long-range dependencies has led to its widespread adoption in visual tasks. Recently, architectures incorporating Transformer or its combination with CNN have shown competitive results in image fusion~\cite{ma2022swinfusion, wang2022swinfuse, tang2022ydtr, tang2023tccfusion, tang2023datfuse, zhao2023cddfuse}.
Tang et al. introduced YDTR~\cite{tang2022ydtr} that comprehensively maintains local features and significant context information through a Y-shape dynamic Transformer.
Zhao et al. proposed a fusion framework based on CNN-Transformer to enhance fusion performance~\cite{zhao2023cddfuse}.

\subsection{OOD Data Challenges in Real-world}
In practical applications, learning-based models often encounter unforeseen samples that deviate from the training distribution, referred to as OOD data. Consequently, model performance on these OOD data is suboptimal, posing a critical challenge for ensuring the reliability and safety of practical applications~\cite{sehwag2019analyzing, pan2023understanding, yang2023reference}. Various techniques, such as domain generalization~\cite{zhou2022domain} and domain adaptation~\cite{ben2010theory, finn2017model}, have been developed to enable models to adapt to distribution shifts and improve OOD performance. Injecting domain knowledge through data augmentation or self-supervised pre-training can further enhance domain generalization and improve model adaptability. 

Deep learning-based fusion models are increasingly applied in open-world scenarios, where OOD data challenges are becoming prominent~\cite{gawlikowski2023handling}. In practical applications, IVIF models are particularly susceptible to the influence of OOD data due to environmental changes, variations in objects and scenes, and differences in sensors. Despite continuous improvements in the performance of current IVIF models, there is a notable lack of studies addressing the challenges posed by OOD scenarios during model deployment. This oversight contributes to unresolved issues, including potential performance degradation and increased uncertainty in open-world applications.

To address challenges posed by OOD data in real-world scenarios, we propose a multi-view data augmentation, aiming to enhance the robustness and generalization of the model.

\section{Methodology}
In this section, we elaborate on the proposed multi-view data augmentation, detailing the implementation specifics for both external and internal perspectives. Additionally, we introduce the frequency-aware fusion network.

\subsection{Problem Definition}
In practical scenarios, the distribution of test data for model deployment often diverges from that of the training data. In real-world applications, this test distribution is typically unknown and unpredictable, posing significant challenges to model effectiveness. Particularly in IVIF tasks, model performance suffers when confronted with test data that differs from the training distribution. To illustrate the challenges posed by OOD data in infrared-visible fusion, several representative fusion methods are evaluated under varying training and test distributions. As detailed in Table~\ref{tab1}, when the test distribution deviates from the training data, the performance of each existing model degrades to varying degrees. In contrast to these current methods, our proposed approach aims at improving the out-of-distribution performance. This highlights the limitations of current models in handling OOD scenarios, challenging their ability to maintain stable performance in unfamiliar or unpredictable environments.

\subsection{Overview}
The overall framework of our proposed method is illustrated in Fig.~\ref{fig1}. Our approach primarily involves external-view data augmentation to adjust data distribution, and internal-view data augmentation for self-supervised learning. External data augmentation utilizes gamma transformation for channel alignment, reducing distribution shifts between external and target data. This strategy not only effectively expands data samples, but also enhances the generalization capability and adaptability of the model. Internal data augmentation leverages weak-aggressive augmentation and self-supervised loss, enabling the model to learn more generalized feature representations, thereby enhancing robustness. 

\subsection{Multi-view Augmentation}
To better address challenges posed by OOD data in real-world applications, we propose a multi-view data augmentation. For the external view, Top-k Selective Channel Alignment is introduced for auxiliary datasets. 
This augmentation strategy entails applying gamma transformation to the RGB channels to make the external data consistent with the target data, thus effectively increasing the number of samples. For the internal view, weak-aggressive augmentation is employed to establish a self-supervised learning paradigm. Through contrastive learning of different augmented views, our model generates robust fusion results. This multi-view data augmentation strategy enhances the adaptability and generalization of our model to OOD scenarios in real-world applications.

\begin{figure*}[!t]
\centering
\includegraphics[width=0.95\linewidth]{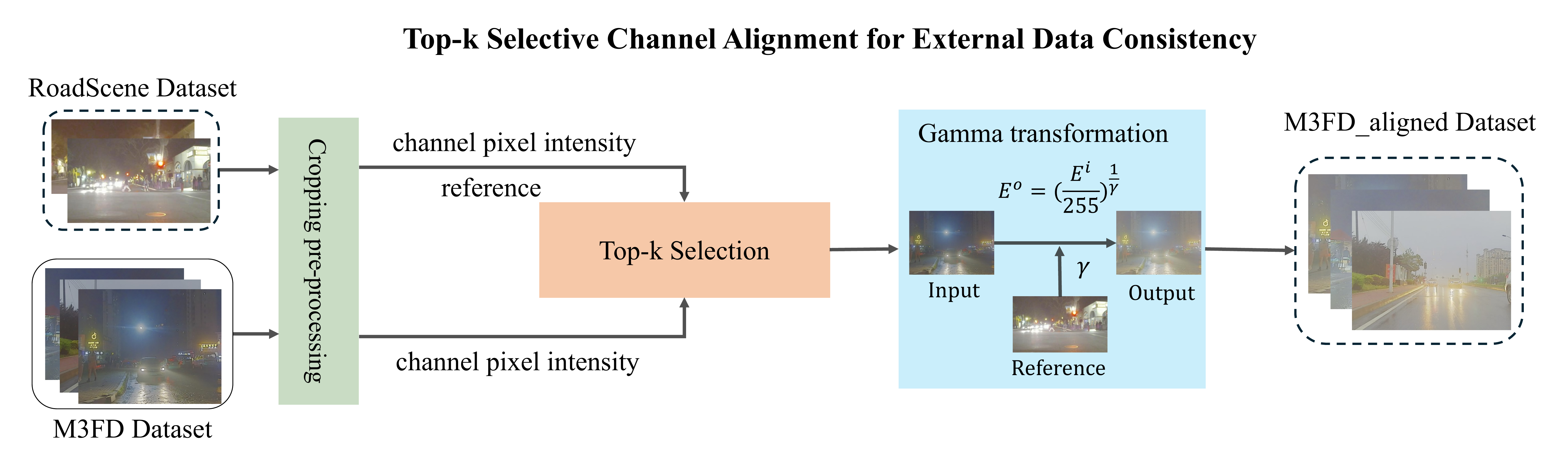}
\caption{Implementation details of external Top-k Selective Channel Alignment. Selectively applying RGB-wise gamma transformation on the localized region of each external image, this external strategy effectively alleviates distribution shifts and achieves data augmentation.}
\label{fig2}
\end{figure*}

\subsubsection{External DA for Auxiliary Dataset}\hspace{0pt} \\
\indent \emph{Top-k Selective Channel Alignment for Data Consistency.} 
To address distribution shifts and enhance data consistency, we propose a novel Top-k Selective Channel Alignment strategy. Unlike traditional augmentation or direct use of auxiliary datasets, this approach harmonizes feature distributions between augmented and target data. As illustrated in Fig.~\ref{fig2}, channel alignment is achieved using gamma transformation at the channel level, which adjusts the pixel distribution of external data to match the target dataset:
\begin{equation}
    E_{R,G,B}^{o} = G(E_{R,G,B}^{i}) = (\frac{E_{R,G,B}^{i}}{255})^{\frac{1}{\gamma_{R,G,B}}}. 
    \label{equa1}
\end{equation}
Here, $G(\cdot)$ represents gamma transformation, $E_{R,G,B}^{i}$ and $E_{R,G,B}^{o}$ denote the input and output pixel intensities of the external data, respectively, ranging from 0 to 255. $\gamma$ represents the gamma factor, and $R$, $G$, and $B$ denote the three channels.

Alignment is applied to localized regions. Both target and external datasets undergo identical cropping operations to facilitate pixel-level consistency and minimize discrepancies across varying image capture conditions (e.g., daytime vs. nighttime). Additionally, a Top-K selection mechanism identifies and transforms only the most relevant patches based on their pixel distribution similarity to the target data. The value of K is determined experimentally to balance augmentation effectiveness and computational cost:
\begin{equation}
    E_{p\_CA} = G(top_{k}(E_{p})). 
    \label{equa2}
\end{equation}
Here, $top_{k}(\cdot)$ denotes the Top-K selection, $E_{p}$ represents the image patches in the auxiliary dataset, and $E_{p\_CA}$ refers to the final aligned external image patches. This approach prioritizes patches with the highest alignment to the target dataset, enhancing data consistency and training efficiency while maintaining robust performance. The overall implementation is shown in Algorithm.~\ref{alg1}.

\begin{algorithm}[!ht]
    \renewcommand{\algorithmicrequire}{\textbf{Input:}}
	\renewcommand{\algorithmicensure}{\textbf{Output:}}
    \caption{Top-k Selective Channel Alignment.}
    \begin{algorithmic}
        \REQUIRE target RoadScene visible images $T$, and external M3FD visible images $E$
        \ENSURE M3FD visible images after channel alignment $E^{CA}$

        \STATE $\triangleright$ Cropping pre-processing
        \STATE Crop RoadScene visible images into $64 \times 64$ patches
        \STATE Crop M3FD visible images into $64 \times 64$ patches

        \STATE $\triangleright$ Calculate channel pixel intensity for RoadScene visible patches

        \FOR{$c$ \textbf{in} $\{R, G, B\}$}
        \STATE \vspace{1mm} $avg\_intensity(\mathbf{T}_c) = Mean(\{T_c^n\}_{n_1=1}^{N_1})$
        \vspace{1mm}
        \ENDFOR

        \STATE $\triangleright$ Top-k selection on M3FD patches
        \FOR{$n_2$ from 1 to $N_2$}
        \STATE Calculate average intensity and difference for $E^{n_2}$
            \FOR{$c$ \textbf{in} $\{R, G, B\}$}
                \STATE \vspace{1mm} $avg\_intensity(\mathbf{E}_c^{n_2}) = Mean(R_c^{n_2})$
            \vspace{1mm}
            \ENDFOR
            \STATE \vspace{1mm} $diff(E^{n_2}, T) = \sum_{c \in \{R, G, B\}} \lvert avg\_intensity(\mathbf{E}_c^{n_2}) - avg\_intensity(\mathbf{T}_c) \rvert$
            \STATE \vspace{1mm} Append $diff(E^{n_2}, T)$ to $diff\_values$
        \vspace{1mm}
        \ENDFOR
        \STATE Sort $diff\_values$ and find indices of top $K$ values
        \STATE Select patches corresponding to these indices and store them in $top\_k\_patches$
        \STATE Return $top\_k\_patches$

        \STATE $\triangleright$ Calculate channel pixel intensity for Top-K M3FD visible patches
        \FOR{$c$ \textbf{in} $\{R, G, B\}$}
            \STATE \vspace{1mm} $avg\_intensity(\mathbf{E}_c) = Mean(\{E_c^k\}_{k=1}^{K})$
        \vspace{1mm}
        \ENDFOR

        \STATE $\triangleright$ Get $E^{CA}$
        \STATE Calculate gamma factor $\gamma_R$, $\gamma_G$, $\gamma_B$ for each channel
        \FOR{$k$ from 1 to $K$}
            \STATE \vspace{1mm} $E_R^{CA} = {(\frac{\mathbf{E}_R}{255})}^{\frac{1}{\gamma_R}}$
            \STATE \vspace{1mm} $E_G^{CA} = {(\frac{\mathbf{E}_G}{255})}^{\frac{1}{\gamma_G}}$
            \STATE \vspace{1mm} $E_B^{CA} = {(\frac{\mathbf{E}_B}{255})}^{\frac{1}{\gamma_B}}$
        \ENDFOR
        \STATE Then we can get the final M3FD visible data after channel alignment.
    \end{algorithmic}
\label{alg1}
\end{algorithm}

\begin{figure*}[!t]
\centering
\includegraphics[width=1\linewidth]{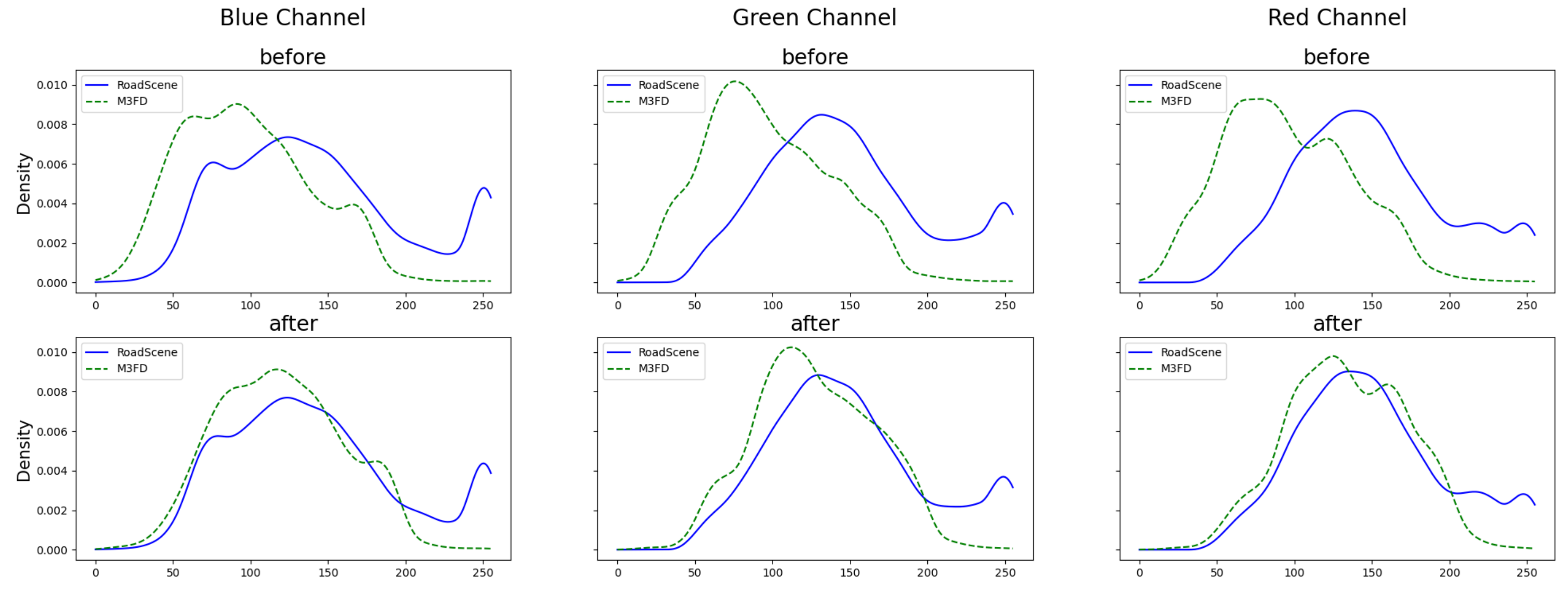}
\caption{Comparison of the RGB-wise distribution of datasets before and after Top-K channel alignment processing. Following channel alignment, the distribution of the external-augmented dataset closely approximates that of the target dataset.}
\label{fig3}
\end{figure*}

\begin{figure}[t]
\centering
\includegraphics[width=1\columnwidth]{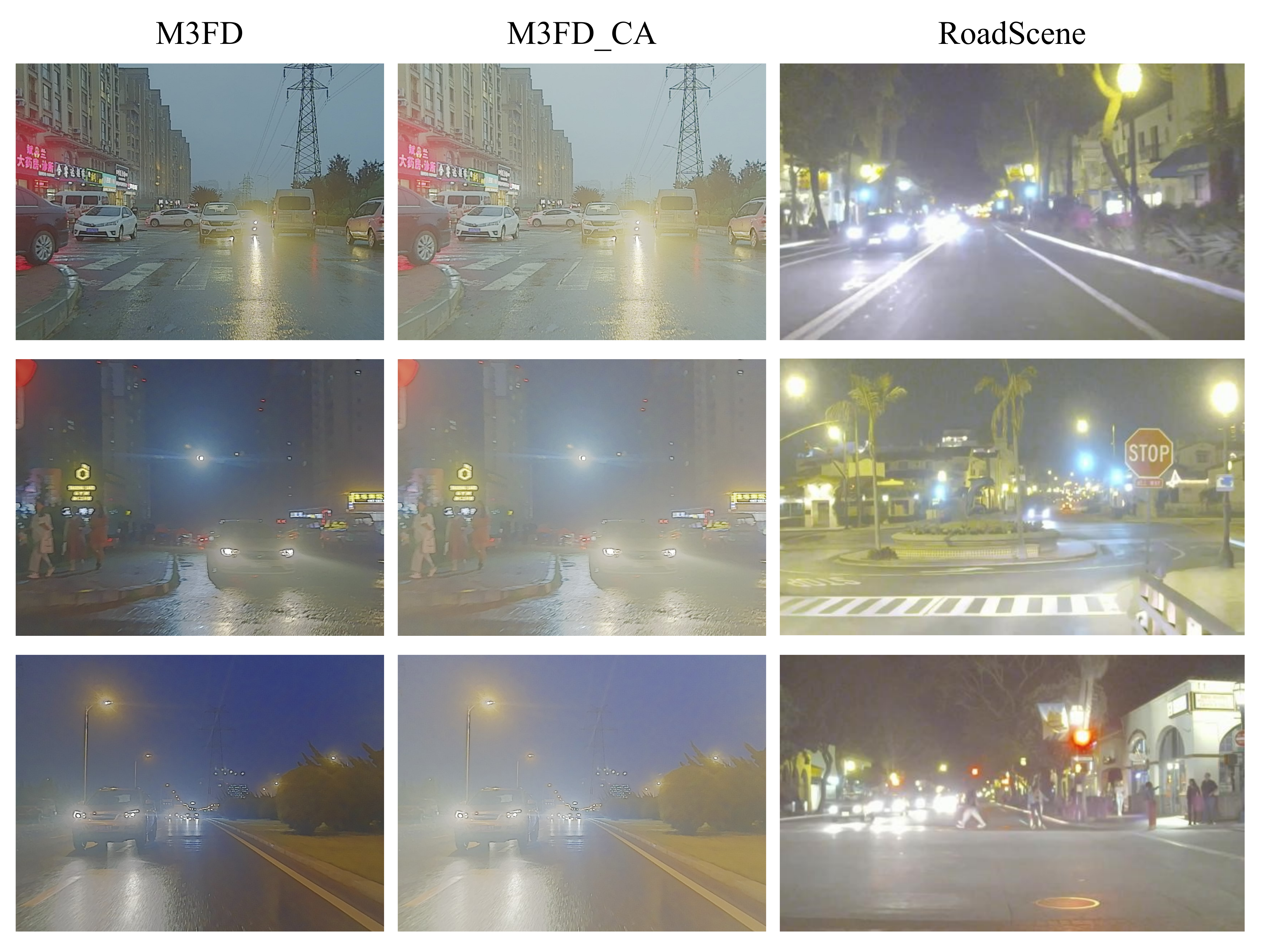}
\caption{Example images from M3FD and RoadScene datasets to show the effects of channel alignment.}
\label{fig4}
\end{figure}

\begin{figure}[!t]
\centering
\includegraphics[width=1\linewidth]{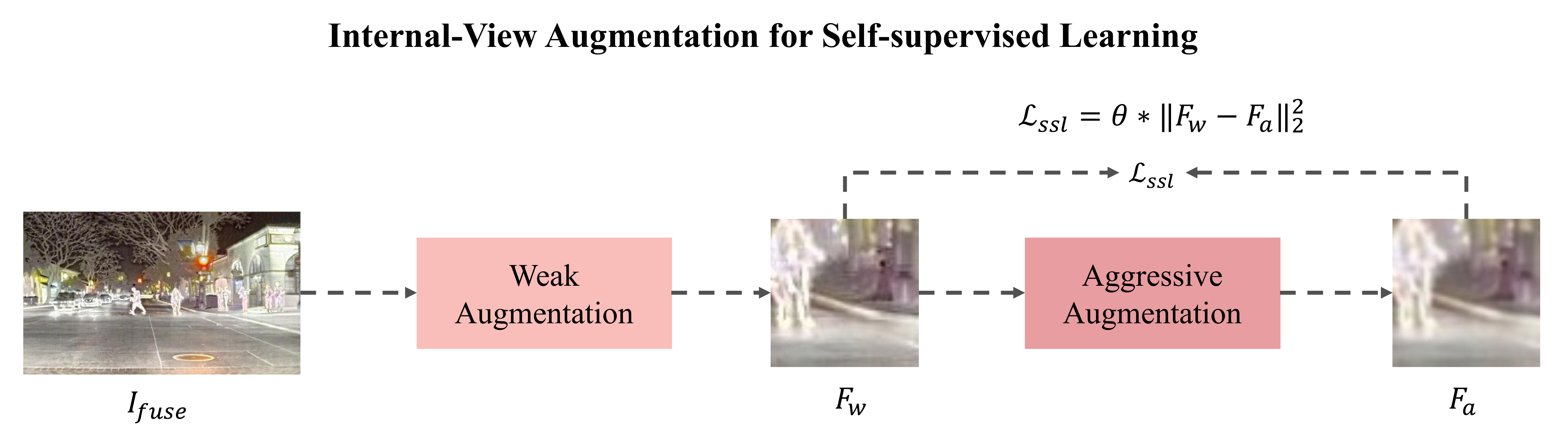}
\caption{The implementation details of the Internal-view Augmentation for Self-supervised Learning. Through contrasting multiple augmented views through weak-aggressive augmentation, self-supervised learning is established, enhancing the final fusion outcomes.}
\label{fig5}
\end{figure}

As depicted in Fig.~\ref{fig3}, the changes in the histograms highlight that there is a significant difference in the color distribution between the original external dataset and the target dataset. Following channel alignment, the pixel distribution in each channel of the external dataset closely resembles that of the target data, thus enhancing the consistency of the external data.

Fig.~\ref{fig4} illustrates the visual impact of channel alignment on the external images. The results clearly show that after implementing channel alignment, the external M3FD~\cite{liu2022target} images are visually closely aligned with the target RoadScene~\cite{xu2020u2fusion} images visually.

\subsubsection{Internal DA for Self-supervised Learning}\hspace{0pt} \\
\indent \emph{Internal-view Augmentation for Self-supervised Learning.} To boost the capability of the model to  generalize across domains and improve fusion quality, we adopt weak-aggressive augmentation for self-supervised learning from an internal data perspective. As shown in Fig.~\ref{fig5}, progressive weak-aggressive augmentation generates correlated multi-level augmented views, which are utilized by the model for contrastive learning. Initially, weak augmentations are applied to produce the corresponding weak-augmented views for fusion results:
\begin{equation}
    F_{w} = weak(I_{fuse}). 
    \label{equa3}
\end{equation}
Here, $weak(\cdot)$ denotes the weak augmentation operation, which is specifically implemented as random cropping. $I_{fuse}$ represents the output of the fusion network, and $F_{w}$ is the corresponding weak-augmented view. Subsequently, aggressive augmentations are carried out on $F_{w}$ to generate the corresponding aggressive-augmented view:
\begin{equation}
    F_{a} = aggr(F_{w}). 
    \label{equa4}
\end{equation}
Here, $aggr(\cdot)$ represents the aggressive augmentation operation, specifically implemented as Gaussian blur. $F_{a}$ denotes the aggressive-augmented view derived from $F_{w}$.

By contrasting these different augmented views, the model can learn more refined feature representations. Specifically, weak augmentations apply basic transformations that help the model capture essential features of the data, while aggressive augmentations introduce more pronounced variations, compelling the model to maintain feature consistency. This approach encourages the model to learn robust and invariant features that are less sensitive to data perturbations, thereby enhancing feature discrimination and generalization. Consequently, a self-supervised loss function is designed to evaluate the consistency between the two augmented views (see Section~\ref{subsec:loss_function} for details). The overall implementation is depicted in Algorithm.~\ref{alg2}.

\begin{algorithm}[t]
    \renewcommand{\algorithmicrequire}{\textbf{Input:}}
    \renewcommand{\algorithmicensure}{\textbf{Output:}}
    \caption{Internal DA for Self-Supervised Learning.}
    \begin{algorithmic}
        \REQUIRE fusion results $I_{fuse}$
        \ENSURE multi-augmentation views $F_w$, $F_a$, and learning loss $\mathcal{L}_{ssl}$
        \FOR{each fusion result $I_{fuse}$}
            \STATE $weak() = Crop()$
            \STATE $aggr() = Gaussian Blur()$
            \STATE Generate weak-augmentation view: $F_{w} = weak(I_{fuse})$
            \STATE Generate aggressive-augmentation view: $F_{a} = aggr(F_{w})$
            \STATE Calculate learning loss: $\mathcal{L}_{ssl} = \theta * \|F_w - F_a\|_2^2 $
        \ENDFOR
    \end{algorithmic}
    \label{alg2}
\end{algorithm}

\subsection{Frequency-aware Fusion Network}
We construct a frequency-aware fusion network to extract and integrate global information and local features across multiple scales. As illustrated in Fig.~\ref{fig1}, the fusion network consists of multi-scale feature extraction and fusion. This structure enables the fused image to preserve comprehensive global structural information and local texture details from the frequency perspective.

\emph{Multi-scale Feature Extraction-Fusion.} Typically, the co-occurrence features of modalities, such as general background and large-scale environmental characteristics, are primarily reflected in the low-frequency domain, whereas high-frequency information emphasizes detailed features within each modality. The proposed fusion network enriches the information within the fused image and enhances the representation of scene targets by extracting and integrating multi-scale modality features.

\begin{figure}[t]
\centering
\includegraphics[width=0.98\columnwidth]{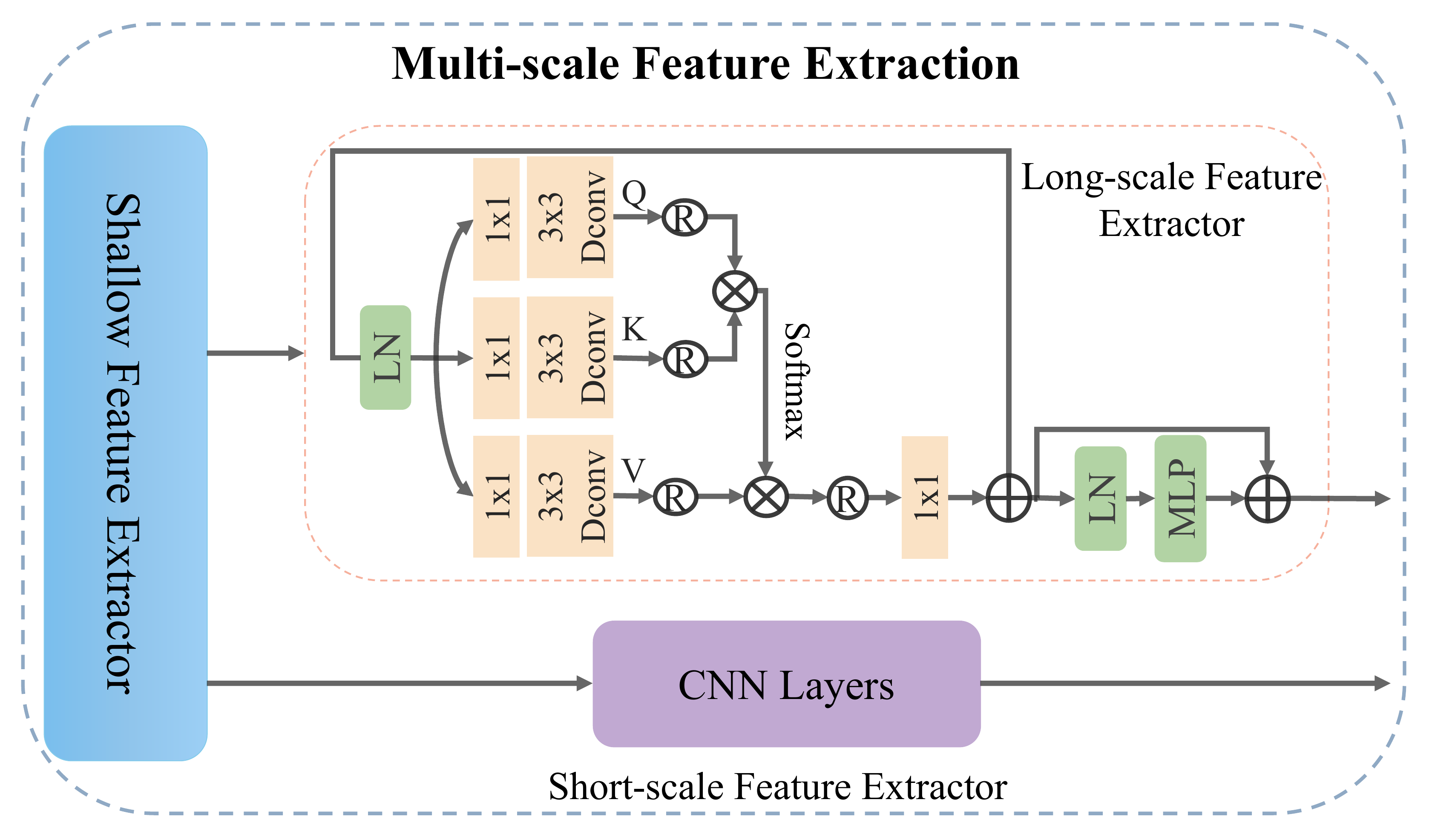}
\caption{The implementation details of the multi-scale feature extraction.}
\label{fig6}
\end{figure}

As shown in Fig.~\ref{fig6}, distinct feature extractors are tailored to extract varied feature representations from the source images. Initially, shallow features are derived from the infrared and visible inputs:
\begin{equation}
    X_{I_{ir}}^{S} = S(I_{ir}), X_{I_{vis}}^{S} = S(I_{vis}),
    \label{equa5}
\end{equation}
where $S(\cdot)$ denotes the shallow feature extractor composed of Restormer blocks~\cite{zamir2022restormer}. $X_{I_{ir}}^{S}$ and $X_{I_{vis}}^{S}$ represent the shallow features extracted from the infrared and visible images, respectively.

Subsequently, low-frequency global features are extracted from these shallow features:
\begin{equation}
    X_{I_{ir}}^{L} = L(X_{I_{ir}}^{S}), X_{I_{vis}}^{L} = L(X_{I_{vis}}^{S}).
    \label{equa6}
\end{equation}
Here, $L(\cdot)$ refers to the long-scale feature extractor composed of Transformer blocks, adept at capturing long-range dependencies. Concurrently, high-frequency detail features are extracted from the shallow features:
\begin{equation}
    X_{I_{ir}}^{H} = H(X_{I_{ir}}^{S}), X_{I_{vis}}^{H} = H(X_{I_{vis}}^{S}).
    \label{equa7}
\end{equation}
Here, $H(\cdot)$ represents the short-scale feature extractor, consisting of convolutional blocks including 1$\times$1 convolutions, 3$\times$3 depth-wise separable convolutions, and ReLU activation functions, to capture abundant local features.

The low-frequency and high-frequency features from each modality are concatenated along the channel dimension and then reconstructed through a reconstruction block:
\begin{equation}
    I_{ir}^{'} = D(X_{I_{ir}}^{L}, X_{ir}^{H}), I_{vis}^{'} = D(X_{I_{vis}}^{L}, X_{vis}^{H}).
    \label{equa8}
\end{equation}
Here, $D(\cdot)$ denotes the reconstruction block, built upon Restormer blocks~\cite{zamir2022restormer}.

To enhance the clarity and detail of the fused image, we employ two distinct fusion blocks to integrate modality features at different scales:
\begin{equation}
    X^{L} = F_{L}(X_{I_{ir}}^{L}, X_{I_{vis}}^{L}), X^{H} = F_{H}(X_{I_{ir}}^{H}, X_{I_{vis}}^{H}).
    \label{equa9}
\end{equation}
Here, $F_{L}$ and $F_{H}$ represent the low-frequency and high-frequency fusion modules, respectively. The structures of these fusion blocks are consistent with the long-scale and short-scale feature extractors. Finally, the low-frequency and high-frequency fused features are concatenated along the channel dimension and input into the reconstruction block to generate the final fused image:
\begin{equation}
    I_{fuse} = D(X^{L}, X^{H}).
    \label{equa10}
\end{equation}

\subsection{Loss Function}
\label{subsec:loss_function}
We adopt a two-stage training strategy, with the loss function comprising feature reconstruction loss and feature fusion loss.

During feature extraction and reconstruction, it is crucial to preserve complete information in the images. Thus, the total loss function in the first stage is defined as follows:
\begin{equation}
    \mathcal{L}_{total}^{rec} = \mathcal{L}_{ir\_rec} + \mathcal{L}_{vis\_rec} + \alpha_{1}\mathcal{L}_{dec}.
    \label{equa11}
\end{equation}
Here, $\mathcal{L}_{ir\_rec}$ and $\mathcal{L}_{vis\_rec}$ represent the reconstruction losses for the infrared and visible images, respectively, while $\mathcal{L}_{dec}$ denotes the feature decomposition loss. The hyperparameter $\alpha_{1}$ balances these three loss components. The reconstruction loss primarily evaluates the similarity between the reconstructed image and the original image:
\begin{equation}
    \mathcal{L}_{ir\_rec} = MSE(I_{ir}, I_{ir}^{'}) + \lambda(1 - SSIM(I_{ir}, I_{ir}^{'})),
    \label{equa12}
\end{equation}
where $MSE$ denotes the mean squared error, and $SSIM$ denotes the structural similarity index, with $\lambda$ as the weighting factor. $\mathcal{L}_{vis\_rec}$ is analogous to $\mathcal{L}_{ir\_rec}$. 
The feature decomposition loss primarily assesses the correlation between features of different frequencies:
\begin{equation}
    \mathcal{L}_{dec} = \frac{(\mathcal{L}_{corr\_H})^2}{\mathcal{L}_{corr\_L}} = \frac{(corr(X_{ir}^{H}, X_{vis}^{H}))^2}{corr(X_{ir}^{L}, X_{vis}^{L}) + \zeta},
    \label{equa13}
\end{equation}
where $corr(\cdot)$ denotes the operation for calculating the correlation coefficient, and $\zeta$ is set to 1.01 to ensure that this term remains positive.

During the fusion stage, the total fusion loss is defined as:
\begin{equation}
    \mathcal{L}_{total}^{fus} = \alpha_{2}\mathcal{L}_{dec} + \alpha_{3}\mathcal{L}_{sim} + \alpha_{4}\mathcal{L}_{ssl}.
    \label{equa14}
\end{equation}
Here, $\mathcal{L}_{sim}$ represents the similarity loss, which evaluates pixel-level and gradient-level discrepancies between the fused result and the source images:
\begin{align}
    \mathcal{L}_{sim} = \frac{1}{HW}\| I_{fuse} - max(I_{ir}, I_{vis}) \|_{1} \notag \\
    + \mu(\frac{1}{HW}\| \lvert \nabla I_{fuse} \rvert -max(\lvert \nabla I_{ir} \rvert, \lvert \nabla I_{vis} \rvert) \|_{1}),
    \label{equa15}
\end{align}

$\mathcal{L}_{ssl}$ is the loss designed for internal DA to measure differences between different augmented views:
\begin{equation}
    \mathcal{L}_{ssl} = \theta * \| F_{w} - F_{a} \|_{2}^{2}, 
    \theta = \theta_{init} * \frac{1}{2}(\cos(\pi\frac{m}{M} + 1)).
    \label{equa16}
\end{equation}
Here, $\theta_{init}$ represents the initial value set at the beginning of training, while $m$ and $M$ denote the current and total training steps. The parameter $\theta$ is key to balancing the contributions of weak and aggressive augmentations in the loss function. Initially, 
$\theta$ is higher, emphasizing aggressive augmentation to learn diverse features. As training progresses, $\theta$ decreases, increasing the influence of weak augmentation, reducing noise impact, and enhancing feature stability and generalization.

\section{Experimental Validation}
In this section, we validate the proposed framework through comprehensive experiments. Firstly, we introduce the experimental setup. Subsequently, we assess the performance of our method both quantitatively and qualitatively by comparing it with the existing state-of-the-art (SOTA) algorithms. Finally, we present several ablation studies to demonstrate the effectiveness of the key modules within our framework.

\subsection{Setup}
\subsubsection{Datasets}
Our experiments are conducted on four publicly available and representative datasets: RoadScene~\cite{xu2020u2fusion}, M3FD~\cite{liu2022target}, MSRS~\cite{tang2022piafusion}, and TNO~\cite{toet2017tno}. When evaluating the performance of current fusion models under varying training and test data distributions, we primarily use the RoadScene dataset (which contains 181 training pairs and 40 test pairs) and the MSRS dataset (which contains 1083 training pairs and 361 test pairs). In implementing the proposed method, we make use of 102 image pairs from the M3FD dataset as the auxiliary dataset, with the RoadScene dataset serving as the target data. Specifically, visible images from both RoadScene and M3FD are randomly cropped into 64$\times$64 patches for Top-K selective channel alignment. We randomly select 181 pairs from RoadScene and combine them with the channel-aligned pairs from M3FD to form our training data.

\subsubsection{Out-of-distribution Setting}
To present an out-of-distribution evaluation, we apply 40 pairs from RoadScene, 361 pairs from MSRS, and 25 pairs from TNO as test sets. The proposed method and other comparative algorithms are evaluated on these datasets. Among thee test sets available, the TNO test set serves as an independent one utilized for OOD evaluation. Additionally, all ablation studies are carried out using the TNO dataset as well.

\subsubsection{Metrics}
We use nine metrics to comprehensively evaluate the fusion results: entropy (EN), mutual information (MI), standard deviation (SD), spatial frequency (SF), average gradient (AG), visual information fidelity (VIF), sum of the correlations of differences (SCD), Qabf, and structural similarity index measure (SSIM).

\subsubsection{Implement details}
All training samples are randomly cropped into patches of size 64$\times$64. The Adam optimizer is used to optimize network parameters over 160 epochs, with the first and second stages consisting of 40 and 120 epochs, respectively. The initial learning rate is set to 1e-4 and decreases by 0.5 every 20 epochs. 
The hyperparameters $\alpha_{1}$, $\alpha_{2}$, $\alpha_{3}$, and $\alpha_{4}$ in the loss functions Eq.(\ref{equa11}) and Eq.(\ref{equa14}) are set to 2, 2, 1, and 1, respectively, to balance the contributions of each loss term and maintain stable optimization. The hyperparameter $\mu$ in the Eq.(\ref{equa14}) is set to 10 to balance intensity and gradient preservation.
The initial value $\theta_{init}$ in the loss term $\mathcal{L}_{ssl}$ is set to 0.1.

\begin{figure*}[!t]
\centering
\includegraphics[width=1\linewidth]{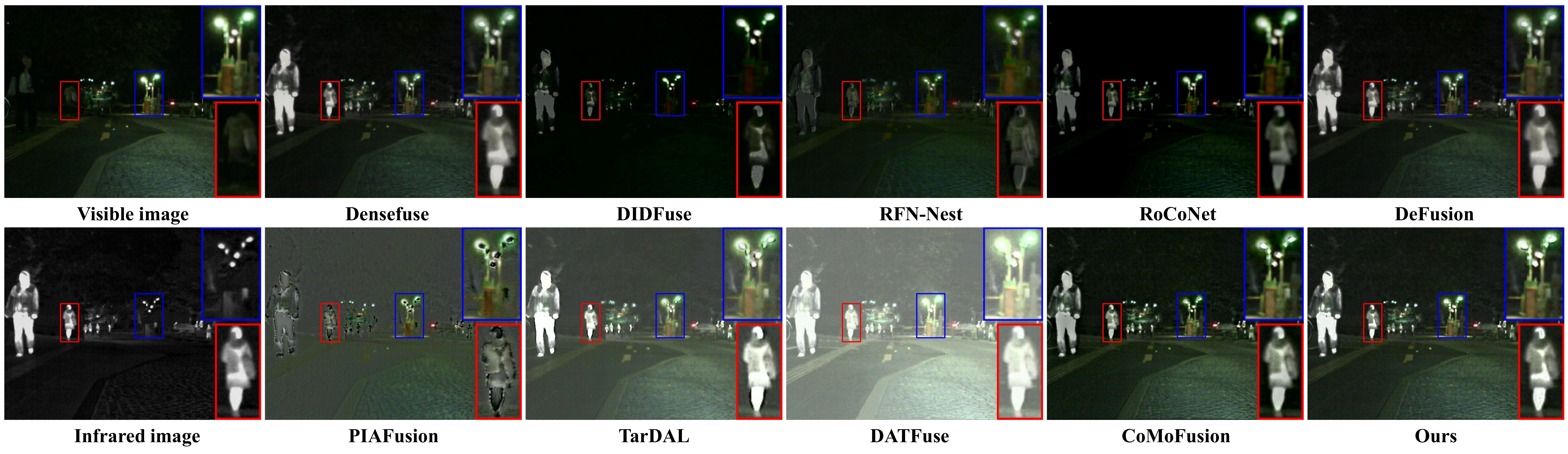}
\caption{Qualitative comparison for “00894N” in MSRS dataset.}
\label{fig7}
\end{figure*}

\begin{figure*}[!t]
\centering
\includegraphics[width=1\linewidth]{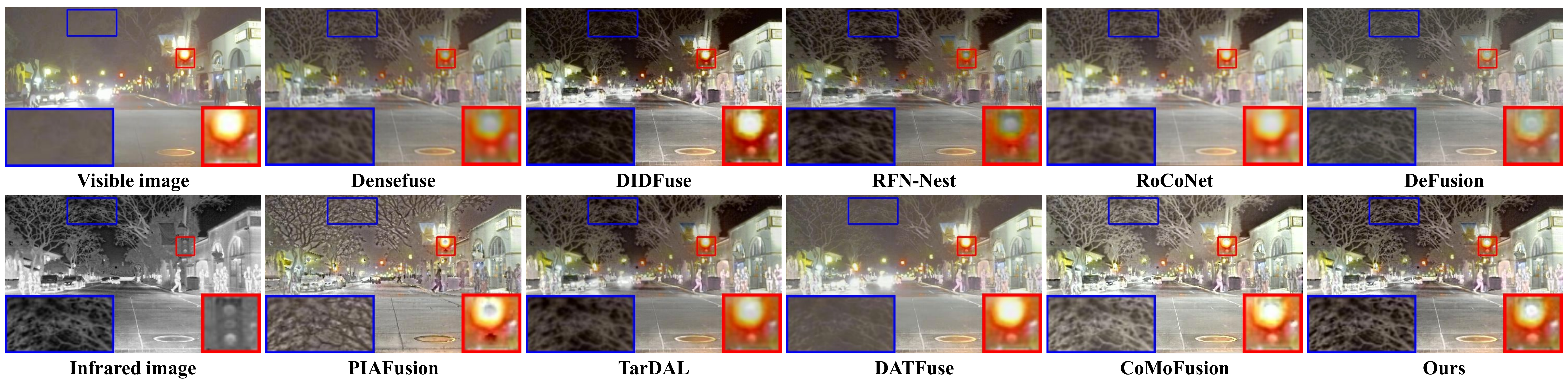}
\caption{Qualitative comparison for "FLIR\_09437" in RoadScene dataset.}
\label{fig8}
\end{figure*}

\begin{figure*}[!t]
\centering
\includegraphics[width=1\linewidth]{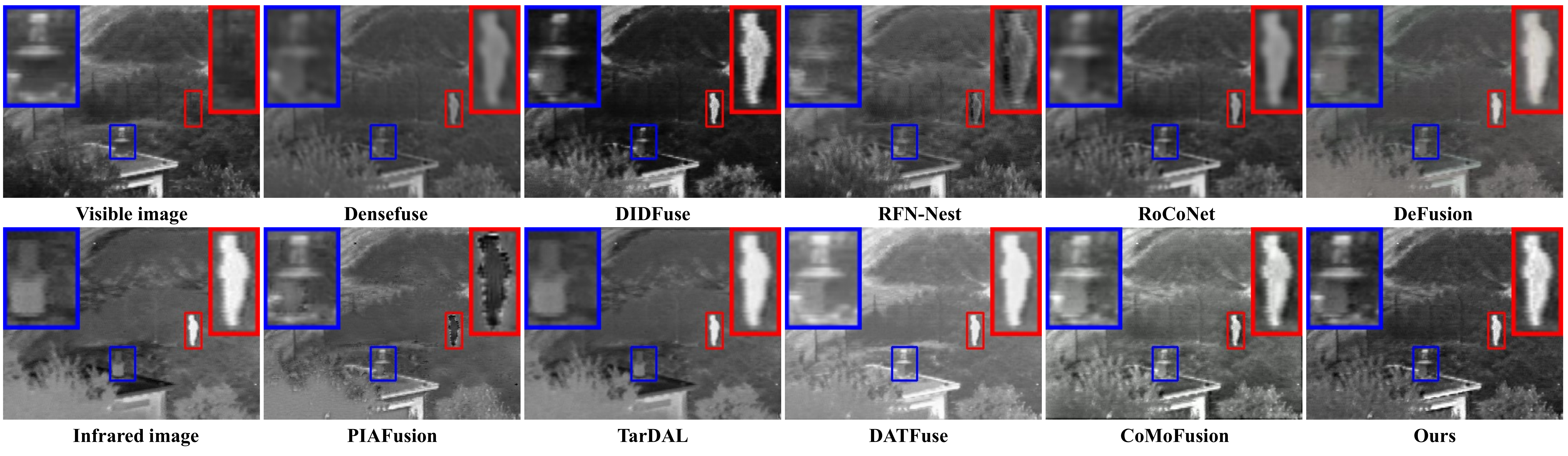}
\caption{Qualitative comparison for “19” in TNO dataset.}
\label{fig9}
\end{figure*}

\subsection{Comparison with SOTA methods}
Our method is compared with nine SOTA fusion algorithms, including Densefuse~\cite{li2018densefuse}, DIDFuse~\cite{ZhaoDIDFuse2020}, RFN-Nest~\cite{li2021rfn}, ReCoNet~\cite{huang2022reconet}, DeFusion~\cite{liang2022fusion}, PIAFusion~\cite{tang2022piafusion}, TarDAL~\cite{liu2022target}, DATFuse~\cite{tang2023datfuse}, and CoMoFusion~\cite{meng2024comofusion}.

\begin{table}[t]
\caption{Quantitative results on the MSRS dataset. Red and blue represent the best and second-best values, respectively.}
\centering
\resizebox{1\columnwidth}{!}{
    \begin{tabular}{c c c c c c c c c c c c}
        \hline
         & Year & EN & MI & SD & SF & AG & VIF & SCD & Qabf & SSIM \\
        \hline
        Densefuse & 2018 & 5.79 & 1.69 & 21.28 & 4.56 & 1.68 & 0.55 & 1.14 & 0.22 & 0.72 \\
        DIDFuse & 2020 & 4.21 & 1.6 & 29.26 & 9.64 & 2.01 & 0.3 & 1.09 & 0.2 & 0.25 \\
        RFN-Nest & 2021 & 5.97 & 1.42 & 23.76 & 5.54 & 1.89 & 0.45 & 1.16 & 0.18 & 0.63 \\
        ReCoNet & 2022 & 4.57 & 1.45 & 33.94 & 6.97 & 2.17 & 0.47 & 1.32 & 0.32 & 0.4 \\
        DeFusion & 2022 & 6.35 & 2.16 & 34.89 & 7.89 & 2.6 & 0.75 & 1.29 & 0.51 & \textcolor{blue}{0.93} \\
        PIAFusion & 2022 & 5.45 & 1.48 & 30.33 & 10.24 & 3.07 & 0.33 & 0.68 & 0.41 & 0.85 \\
        TarDAL & 2022 & 6.35 & 1.82 & 35.48 & 9.91 & 3.12 & 0.67 & 1.48 & 0.42 & 0.71 \\
        DATFuse & 2023 & 6.3 & \textcolor{blue}{2.22} & 31.27 & 8.01 & 2.74 & \textcolor{blue}{0.74} & 1.28 & 0.43 & 0.5 \\
        CoMoFusion & 2024 & \textcolor{red}{6.7} & 2.04 & \textcolor{blue}{41.53} & \textcolor{blue}{11.85} & \textcolor{blue}{3.89} & 0.9 & \textcolor{red}{1.71} & \textcolor{blue}{0.62} & 0.95  \\
        \hline \hline
        \textbf{CrossFuse} & ours & \textcolor{blue}{6.63} & \textcolor{red}{2.29} & \textcolor{red}{44.14} & \textcolor{red}{12.51} & \textcolor{red}{3.96} & \textcolor{red}{0.9} & \textcolor{blue}{1.66} & \textcolor{red}{0.63} & \textcolor{red}{0.95} \\
        \hline
        
    \end{tabular}
}
\label{tab2}
\end{table}

\subsubsection{Qualitative Results}
Fig.~\ref{fig7}, Fig.~\ref{fig8}, and Fig.~\ref{fig9} present a qualitative comparison between our method and several representative fusion algorithms. Our approach effectively combines and retains modality-specific information from the source images, such as the thermal radiation characteristics of infrared images and the rich texture information of visible images.
The fusion results from our method prominently highlight foreground objects in dark areas, enhancing overall scene representation. Moreover, background structures and texture details from visible images exhibit clearer and more distinct edge contours in the fused image. As a result, our method produces fusion results with prominent targets, enhanced contrast, and improved alignment with human visual perception.

As shown in Fig.~\ref{fig7}, we perform a visual comparison on the "00894N" image from the MSRS dataset. The fused images generated by Densefuse, DIDFuse, ReCoNet, PIAFusion, and RFN-Nest methods appear relatively blurred, with targets in the scene (highlighted in red and blue boxes in Fig.~\ref{fig7}) lacking prominence and detailed textures. DeFusion and TarDAL preserve local textures, showing clear target features and detailed scene backgrounds. However, DeFusion tends to favor the infrared image, while TarDAL introduces noise, leading to structural artifacts. The fusion outcomes produced by DATFuse enhance the overall scene brightness at the cost of rich texture details. The fused images yielded by CoMoFusion generally perform well, except that the edge preservation of some targets is less than satisfactory. In contrast, our method produces fused results where the pedestrian (red box) appears clearer and more prominent, and the streetlight (blue box) exhibits more intricate texture details. Overall, our fused images present clear and prominent foreground targets and texture-rich background structures.

As illustrated in Fig.~\ref{fig8}, a visual comparison is conducted on the "FLIR\_09437" scene from the RoadScene dataset. Fusion results generated by the Densefuse, ReCoNet, DeFusion, and RFN-Nest are relatively blurred and fail to highlight infrared targets and distinct background textures (indicated by red and blue boxes in Fig.~\ref{fig8}). DIDFuse preserves more visible texture information but lacks prominent infrared characteristics. PIAFusion introduces excessive noise into the fused images, resulting in unclear scene targets. Although TarDAL achieves satisfactory fusion results to some extent, it still presents artifacts and blurred edges. The outcomes of DATFuse have a deficiency in rich texture details within the background region, and the infrared target particulars are not conspicuous enough. Regarding the results of CoMoFusion, the outlines and edges of the targets lack clarity and sharpness. In contrast, our method better preserves local texture information from visible images and thermal characteristics from infrared images, enhancing the prominence of the traffic light (red box), and improving the clarity of tree contours and textures (blue box).

As shown in Fig.~\ref{fig9}, "19" from the TNO dataset is selected for qualitative comparison. Densefuse, ReCoNet, DeFusion, and RFN-Nest fail to effectively integrate complementary information, resulting in blurred fusion outcomes. DIDFuse preserves more visible texture information and local details, but the fused image generally leans towards the visible image. The fusion result generated by DeFusion fails to highlight the "person" (red box), with unclear contours, and the background area (blue box) lacks detailed textures. TarDAL fails to preserve local textures in visible images, resulting in fusion images that resemble infrared ones more closely. The fusion outcomes of DATFuse and CoMoFusion show that the targets therein are devoid of clear outlines and edges. In contrast, our method achieves a balance between visible and infrared images. Our fused results exhibit clear contours and details for infrared targets, with better-preserved and enhanced background textures. 

\begin{table}[t]
\caption{Quantitative results on the RoadScene dataset. Red and blue represent the best and second-best values, respectively.}
\centering
\resizebox{1\columnwidth}{!}{
    \begin{tabular}{c c c c c c c c c c c}
        \hline
         & Year & EN & MI & SD & SF & AG & VIF & SCD & Qabf & SSIM \\
        \hline
        Densefuse & 2018 & 6.69 & 1.93 & 30.01 & 6.13 & 2.69 & 0.47 & 1.25 & 0.26 & 0.79 \\
        DIDFuse & 2020 & \textcolor{blue}{7.35} & 2.13 & \textcolor{red}{53.22} & 14.49 & 5.6 & 0.58 & \textcolor{red}{1.77} & 0.48 & \textcolor{blue}{0.92} \\
        RFN-Nest & 2021 & 7.15 & 1.95 & 42.28 & 10.99 & 4.24 & 0.41 & 1.47 & 0.29 & 0.75 \\
        ReCoNet & 2022 & 7.01 & 1.94 & 39.35 & 7.95 & 3.33 & 0.49 & 1.57 & 0.28 & 0.81 \\
        DeFusion & 2022 & 6.93 & 2.16 & 36.12 & 8.85 & 3.51 & 0.54 & 1.33 & 0.4 & 0.91 \\
        PIAFusion & 2022 & 7.09 & 1.71 & 40.0 & \textcolor{red}{18.41} & 6.53 & 0.26 & 0.67 & 0.48 & 0.62 \\
        TarDAL & 2022 & 7.2 & 2.4 & 46.29 & 12.06 & 4.59 & 0.53 & 1.33 & 0.43 & 0.87 \\
        DATFuse & 2023 & 7.02 & \textcolor{blue}{2.34} & 43.68 & 11.15 & 4.15 & \textcolor{blue}{0.6} & 1.37 & 0.44 & 0.91 \\
        CoMoFusion & 2024 & 7.17 & 1.46 & 42.32 & 17.15 & \textcolor{red}{7.47} & 0.51 & 1.25 & \textcolor{blue}{0.48} & 0.89 \\
       \hline
        \textbf{CrossFuse} & ours & \textcolor{red}{7.44} & \textcolor{red}{2.41} & \textcolor{blue}{52.83} & \textcolor{blue}{18.31} & \textcolor{blue}{6.71} & \textcolor{red}{0.69} & \textcolor{blue}{1.61} & \textcolor{red}{0.59} & \textcolor{red}{0.96} \\
        \hline
        
    \end{tabular}
}
\label{tab3}
\end{table}

\subsubsection{Quantitative Results}
As presented in Table~\ref{tab2}, Table~\ref{tab3}, and Table~\ref{tab4}, our method undertakes a quantitative comparison with nine SOTA algorithms across the MSRS, RoadScene and TNO datasets. Surveying all the test sets, our method invariably secures a top position, ranking either first or second on every metric. This highlights the robustness and generalization of our approach under diverse conditions and settings, particularly when handling OOD data. The outstanding performance attained across various test distributions powerfully validates that our method can effectively retain texture details and structural edge information. Consequently, our fusion results exhibit enhanced contrast and are in close alignment with human visual perception.

\begin{table}[t]
\caption{Quantitative results on the TNO dataset. Red and blue represent the best and second-best values, respectively.}
\centering
\resizebox{1\columnwidth}{!}{
    \begin{tabular}{c c c c c c c c c c c}
        \hline
         & Year & EN & MI & SD & SF & AG & VIF & SCD & Qabf & SSIM \\
        \hline
        Densefuse & 2018 & 6.26 & 1.42 & 23.39 & 4.14 & 1.87 & 0.45 & 1.42 & 0.21 & 0.75 \\
        DIDFuse & 2020 & 6.74 & 1.58 & \textcolor{blue}{46.68} & 12.5 & 4.32 & 0.59 & 1.7 & 0.37 & 0.79 \\
        RFN-Nest & 2021 & 6.71 & 1.46 & 30.81 & 8.7 & 3.22 & 0.4 & 1.42 & 0.21 & 0.73 \\
        ReCoNet & 2022 & 6.77 & 1.53 & 37.72 & 6.65 & 2.67 & 0.5 & \textcolor{red}{1.72} & 0.29 & 0.76 \\
        DeFusion & 2022 & 6.58 & 1.76 & 30.99 & 6.6 & 2.67 & 0.55 & 1.51 & 0.36 & 0.94 \\
        PIAFusion & 2022 & 6.73 & 1.74 & 35.33 & 12.29 & 4.43 & 0.38 & 1.08 & 0.45 & 0.96 \\
        TarDAL & 2022 & 6.78 & 1.95 & 40.6 & 11.5 & 4.15 & 0.57 & 1.5 & 0.41 & 0.94 \\
        DATFuse & 2023 & 6.79 & \textcolor{red}{2.25} & 38.13 & 8.45 & 3.45 & 0.63 & 1.47 & 0.42 & 0.82 \\
        CoMoFusion & 2024 &  \textcolor{blue}{7.1} & 1.44 & 40.89 & \textcolor{red}{15.13} & \textcolor{blue}{5.37} & \textcolor{blue}{0.64} & 1.58 & \textcolor{blue}{0.49} & \textcolor{blue}{0.98}  \\
        \hline \hline
        \textbf{CrossFuse} & ours & \textcolor{red}{7.21} & \textcolor{blue}{2.13} & \textcolor{red}{49.11} & \textcolor{blue}{14.16} & \textcolor{red}{5.49} & \textcolor{red}{0.77} & \textcolor{blue}{1.7} & \textcolor{red}{0.54} & \textcolor{red}{0.99} \\
        \hline
        
    \end{tabular}
}
\label{tab4}
\end{table}

\subsubsection{Efficiency Results}
In order to conduct a more exhaustive assessment of model performance, we made a comparison between the efficiency of the proposed framework and that of nine comparative algorithms, principally concentrating on the parameter size, floating point operations (FLOPs), and model inference time. As depicted in Table~\ref{tab5}, our method displays a lower running efficiency compared with Densefuse, TarDAL and CoMoFusion. With the aim of generating fused results that are as information-rich and of as high quality as possible, our method utilizes the Transformer structure and incorporates multiple components, inevitably forfeiting a certain degree of inference efficiency. Nevertheless, on benchmark datasets with diverse distributions, our method exhibits outstanding performance. How to further strike a balance between performance and efficiency will also be the central focus of our follow-up work.

\begin{table}[t]
\caption{Efficiency comparisons of our method with nine state-of-the-art algorithms. Black bold represents the best results.}
\centering
\resizebox{1\columnwidth}{!}{
    \begin{tabular}{c c c c c c c}
        \hline
        & & & MSRS & RoadScene & TNO \\
        \hline
        Algorithm & Size(M) & FLOPs(G) & & Running time(s) & \\
        Densefuse & \textbf{0.074} & 4.458 & \textbf{0.0017} & \textbf{0.0018} & \textbf{0.0064} \\
        DIDfuse & 4285.60 & - & 0.0113 & 0.0576 & 0.0878\\
        RFN-Nest & 10.93 & 520.85 & 0.0760 & 0.0386 & 0.0746\\
        ReCoNet & 1072.74 & 12.55 & 0.3782 & 0.1602 & 0.3227\\
        DeFusion & 57.64 & 15.17 & 0.0617 & 0.0401 & 0.0645\\
        PIAFusion & 4813.23 & \textbf{3.21} & 0.039 & 0.0229 & 0.0358\\
        TarDAL & 0.296 & 14.88 & 0.0073 & 0.0422 & 0.0544\\
        DATFuse & 260.35 & - & 0.0474 & 0.0203 & 0.0443\\
        CoMoFusion & 4.69 & 17.99 & 0.006 & 0.0148 & 0.0091\\
        \hline \hline
        \textbf{CrossFuse} (Ours) & 0.80 & 3.30 & 0.0121 & 0.0188 & 0.019  \\
        \hline
        
    \end{tabular}
}
\label{tab5}
\end{table}


\begin{table}[t]
\caption{Ablation experiment results in the test set of TNO. Red indicates the best value, and blue indicates the second-best value.}
\centering
\resizebox{1\columnwidth}{!}{
    \begin{tabular}{c c c c c c c}
        \hline
         & EN & MI & SCD & VIF & Qabf & SSIM \\
        \hline
        Multi-View Data Augmentation\\
        \hline
        I.w/o Top-k Selective Channel Alignment & 7.2 & 2.19 & 1.67 & 0.77 & 0.55 & 0.99 \\
        II.w/o Internal-view DA ($\mathcal{L}_{ssl}$) & 7.22 & 2.08 & 1.69 & 0.75 & 0.54 & 0.96 \\
        \hline
        Frequency-aware Fusion Network\\
        \hline
        III.CNN Layers $\rightarrow$ Transformer Blocks in $L(\cdot)$ & 7.19 & 2.0 & 1.61 & 0.72 & 0.53 &0.96 \\
        IV.Transformer Blocks $\rightarrow$ CNN Layers in $H(\cdot)$ & 7.14 & 2.08 & 1.66 & 0.75 & 0.54 & 0.97 \\ \hline \hline
        \textbf{Ours (Full model)} & \textcolor{blue}{7.21} & \textcolor{blue}{2.13} & \textcolor{red}{1.7} & \textcolor{red}{0.77} & \textcolor{blue}{0.54} & \textcolor{red}{0.99} \\
        \hline
    \end{tabular}
}
\label{tab6}
\end{table}

\subsection{Ablation Study}
We perform ablation experiments to evaluate the effectiveness of various modules within our proposed framework, focusing on their contribution to the overall fusion performance. To quantitatively assess the performance, we employ metrics such as MI, SCD, VIF, Qabf, and SSIM.

\subsubsection{Top-k Selective Channel Alignment}

In the first set of ablation experiments, we compare the performance of models with and without the Top-k Selective Channel Alignment strategy. In the baseline setting, we merged two datasets without any selective alignment. In contrast, in the comparison setting, we applied the Top-k alignment. As detailed in Table~\ref{tab6}, the improved EN and SCD metrics indicate that the alignment strategy enhances data consistency and model performance. The improvement in performance is due to the fact that this alignment makes the model focus on relevant features, thereby improving fusion consistency and overall quality.

\subsubsection{Internal-view DA for self-supervised learning}
We assess the impact of internal data augmentation by comparing models with and without weak-aggressive augmentations and the $\mathcal{L}{ssl}$ term in the loss function Eq.(\ref{equa14}). As demonstrated in Table~\ref{tab6}, removing the $\mathcal{L}{ssl}$ term results in a significant decline in performance, which is manifested by lower MI and SSIM values. This emphasizes that internal data augmentation assists the model in retaining richer feature representations across augmented views, thereby leading to better contrast and visual quality. The improvement stems from the diversity and consistency brought about by internal augmentation, which bolsters the ability of the model to generalize.

\subsubsection{Multi-scale Feature Extraction \& Fusion}
In the ablation study of the frequency-aware fusion network, we explore the multi-scale feature extraction and fusion strategy. Specifically, we replace the Transformer block with a CNN layer in the long-scale extractor and vice versa in the short-scale extractor, and do the opposite (replace the CNN layer with a Transformer block) in the short-scale extractor. The results shown in Table~\ref{tab6} indicate that replacing either component would lead to a decline in all metrics, which verifies that both elements are indispensable for maintaining image details and overall performance. The performance degradation implies that multi-scale fusion plays a crucial role in preserving both detailed and global information, contributing to the generation of fused images that are clearer and possess more detailed information.

\emph{Limitation:} we design an infrared-visible image fusion framework tackles out-of-distribution (OOD) scenarios in real-world environments by adopting a data-centric strategy. Through multi-view data augmentation techniques, this strategy bolsters the adaptability and generalization of the model to diverse data. Notwithstanding the outstanding performance of our framework under various experimental conditions, several limitations persist. Firstly, the model performance might deteriorate in the face of extreme distribution shifts, especially when there is a conspicuous disparity between the auxiliary and target datasets. Finally, while the framework fares well within the experimental setup, further research is requisite to widen its applicability to more complex scenarios or other tasks. Nevertheless, the proposed framework proffers a novel means of confronting OOD challenges in infrared-visible image fusion, providing a foundation for future work aimed at improving the robustness and adaptability of fusion models in real-world applications. We anticipate that this framework will kindle further progress in deploying fusion models in dynamic, real-world scenarios within the image fusion community.

\section{Conclusion and Future Work}
In this paper, we delve into the challenges posed by OOD data in infrared-visible fusion for real-world applications. We put forward an efficient image fusion framework based on multi-view data augmentation,which incorporates Top-K selective channel alignment and weak-aggressive augmentation techniques for both external-view and internal-view data augmentation. This strategy augments the generalization and robustness of our model, enabling it to adeptly handle the OOD data challenges commonly encountered in practical scenarios. Extensive experiments corroborate that our framework attains superior fusion performance across diverse data distributions, substantially propelling infrared-visible fusion for real-world applications. In future research, we intend to concentrate on exploring more efficacious methods to further grapple with OOD challenges, enhance model performance, and augment its deployment in real-world scenarios. Beyond image fusion, the core tenets of cross-domain alignment and multi-view augmentation hold the promise of benefiting other tasks, such as object detection, semantic segmentation, and image super-resolution. These tasks could also profit from enhanced data consistency and improved cross-domain learning. Our future endeavors will involve adapting and extending our approach to optimize performance for these tasks while contending with their specific challenges.

\bibliographystyle{IEEEtran}
\bibliography{ref}

\begin{IEEEbiography}
[{\includegraphics[width=1in,height=1.25in,clip,keepaspectratio]{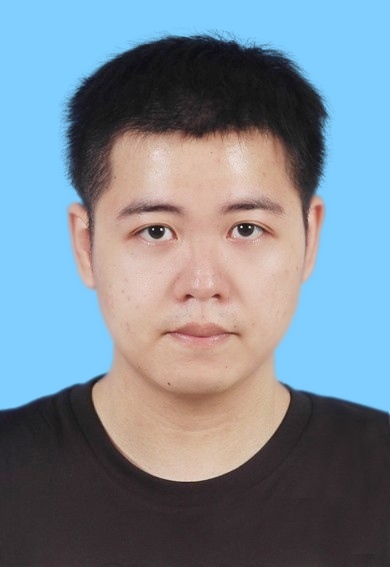}}]{Yukai Shi}
received the Ph.D. degrees from the school of Data and Computer Science, Sun Yat-sen University, Guangzhou China, in 2019. He is currently a lecturer at the School of Information Engineering, Guangdong University of Technology, China. His research interests include computer vision and machine learning.
\end{IEEEbiography}

\begin{IEEEbiography}[{\includegraphics[width=1in,height=1.25in,clip,keepaspectratio]{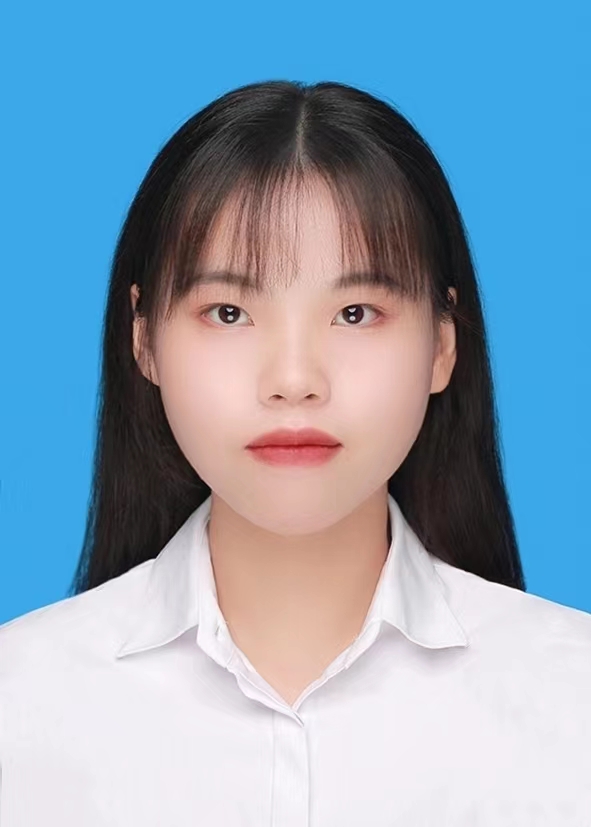}}]{Cidan Shi} received the B.S. degree in 2022, from the School of Information Engineering, Guangdong University of Technology, Guangzhou, China, where she is currently working towards a M.S. degree. Her research interests include computer vision and machine learning.
\end{IEEEbiography}

\begin{IEEEbiography}[{\includegraphics[width=1in,height=1.25in,clip,keepaspectratio]{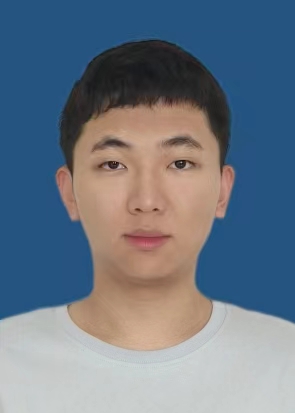}}]{Zhipeng Weng } received the B.S. degree in 2023, from the School of Information Engineering, Guangdong University of Technology, Guangzhou, China, where he is currently working towards a M.S. degree. His research interests include computer vision and machine learning
\end{IEEEbiography}

\begin{IEEEbiography}
[{\includegraphics[width=1in,height=1.25in,clip,keepaspectratio]{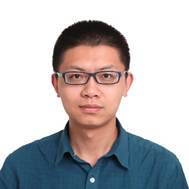}}]{Yin Tian }
received the B.E., M.S., and Ph.D. degrees in transportation engineering from Beijing Jiaotong University, Beijing, China. He is currently a Senior Researcher with the CRRC Academy, Beijing, China. As a Project Manager, he has led the research and development of the CRRC rail prognostics and health management (PHM) system, intelligent wireless sensor networks for rail system, artificial intelligence-based behavior analytic system, etc. His research interests focus on intelligent products and their applications in transportation systems.
\end{IEEEbiography}

\begin{IEEEbiography}
[{\includegraphics[width=1in,height=1.25in,clip,keepaspectratio]{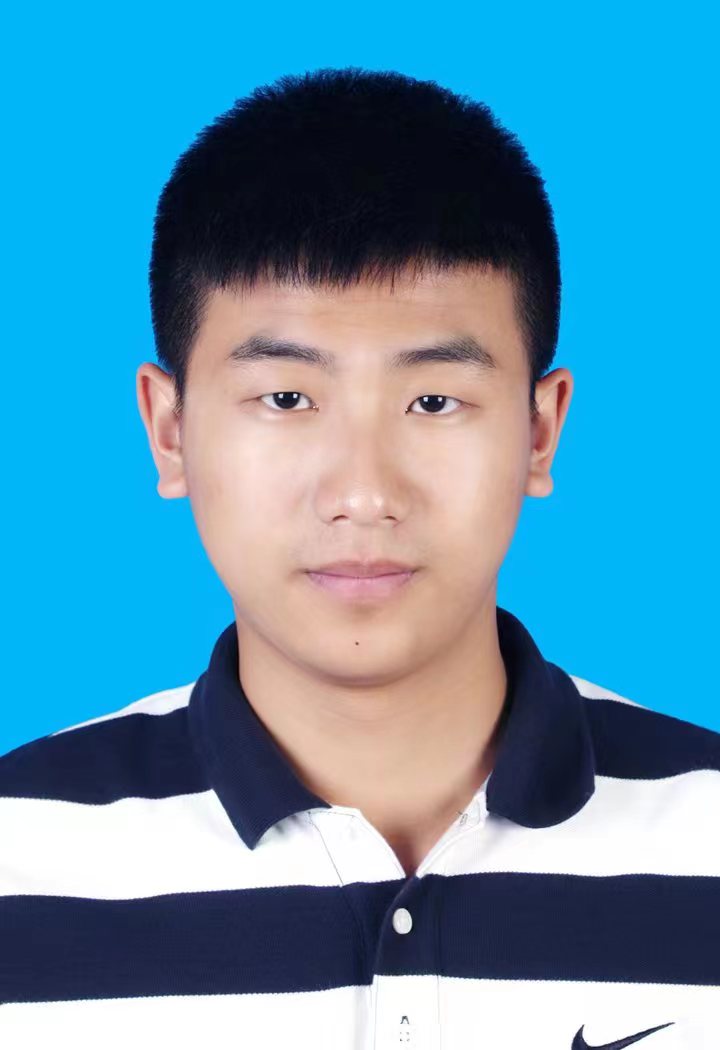}}]{Xiaoyu Xian }
received the M.S. degrees from Beijing Jiaotong University, Beijing, China in 2017. Currently, he is a researcher with the technical department, CRRC Academy Co., Ltd., Beijing. His current research interests include optical character recognition and machine learning.
\end{IEEEbiography}

\begin{IEEEbiography}
[{\includegraphics[width=1in,height=1.25in,clip,keepaspectratio]{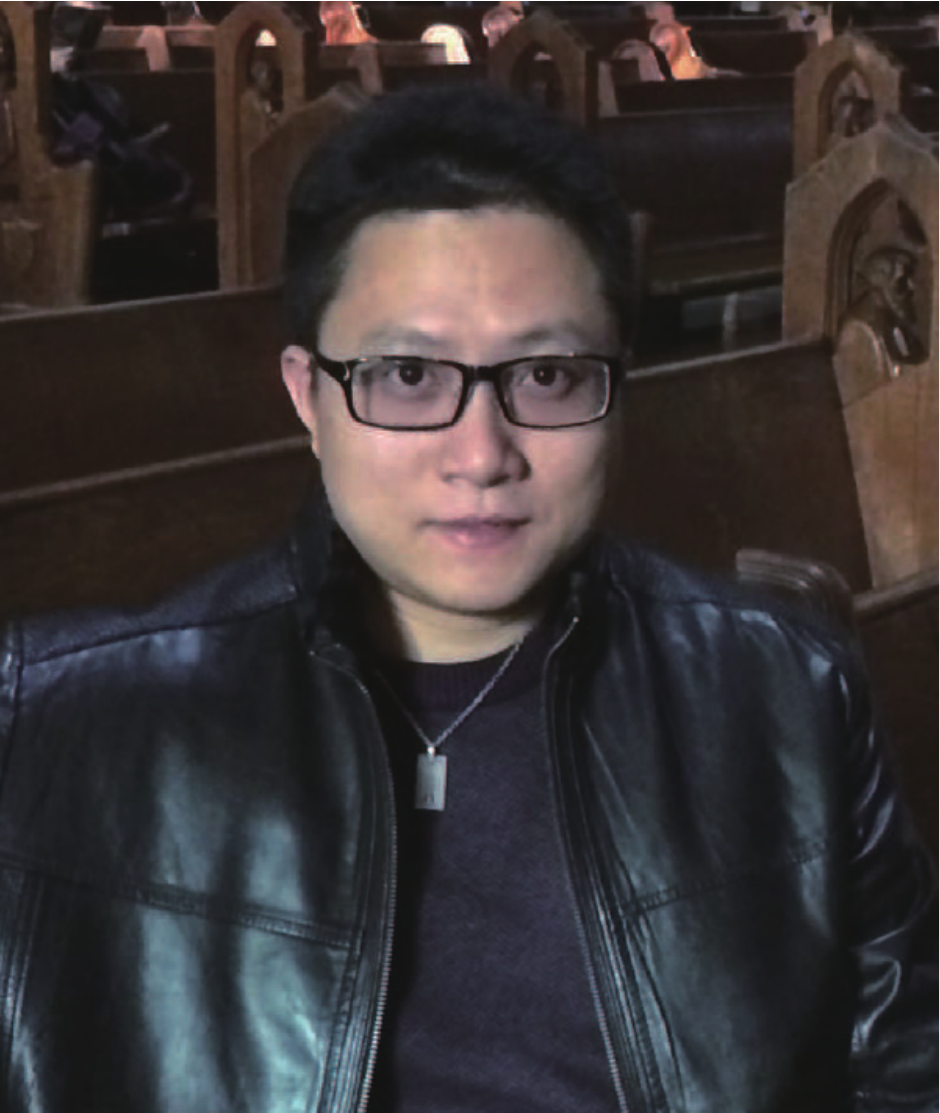}}]{Liang Lin}(Fellow, IEEE) is a Full Professor of computer science at Sun Yat-sen University. He served as the Executive Director and Distinguished Scientist of SenseTime Group from 2016 to 2018, leading the R$\&$D teams for cutting-edge technology transferring. He has authored or co-authored more than 200 papers in leading academic journals and conferences, and his papers have been cited by more than 30,000 times. He is an associate editor of IEEE Trans.Neural Networks and Learning Systems and IEEE Trans. Multimedia, and served as Area Chairs for numerous conferences such as CVPR, ICCV, SIGKDD and AAAI. He is the recipient of numerous awards and honors including Wu Wen-Jun Artificial Intelligence Award, the First Prize of China Society of Image and Graphics, ICCV Best Paper Nomination in 2019, Annual Best Paper Award by Pattern Recognition (Elsevier) in 2018, Best Paper Dimond Award in IEEE ICME 2017, Google Faculty Award in 2012. His supervised PhD students received ACM China Doctoral Dissertation Award, CCF Best Doctoral Dissertation and CAAI Best Doctoral Dissertation. He is a Fellow of IEEE/IAPR/IET.
\end{IEEEbiography}

\vfill

\end{document}